\documentclass[final]{cvpr}

\usepackage{times}
\usepackage{epsfig}
\usepackage{graphics}
\usepackage{float}
\usepackage{subfigure}
\usepackage{amsmath}
\usepackage{amssymb}
\usepackage{mathtools}
\usepackage{algorithm}
\usepackage{algorithmic}
\usepackage[OT1]{fontenc}
\usepackage{balance}
\usepackage{enumerate}
\usepackage{multirow}
\usepackage{bm}
\usepackage{booktabs}
\usepackage[dvipsnames]{xcolor}
\usepackage{eqparbox}

\definecolor{ballblue}{rgb}{0.13, 0.67, 0.8}
\definecolor{lightseagreen}{rgb}{0.13, 0.7, 0.67}
\definecolor{org}{HTML}{F8A145}
\definecolor{blu}{HTML}{63ACE5}
\definecolor{c1}{HTML}{41B3A3}
\definecolor{c2}{HTML}{3500D3}
\definecolor{softred}{HTML}{FE8A71}
\definecolor{softblue}{HTML}{63ACE5}

\usepackage[pagebackref=true,breaklinks=true,colorlinks,bookmarks=false]{hyperref}

\newtheorem{prop}{Proposition}

\newcommand{\first}[1]{\textbf{\color{softred}#1}}
\newcommand{\second}[1]{\textbf{\color{softblue}#1}}

\def \vi {\bm{v}_i}
\def \vhati {\hat{\bm{v}}_i}
\def \vhatj {\hat{\bm{v}}_j}
\def \xii {\bm{x}_i}
\def \xhati {\hat{\bm{x}}_i}
\def \xhatj {\hat{\bm{x}}_j}
\def \yi {{y}_i}
\def \yj {{y}_j}
\def \wyi {\bm{\omega}_{\yi}}
\def \wj {\bm{\omega}_{j}}

\def\cvprPaperID{2586} % *** Enter the CVPR Paper ID here
\def\confYear{CVPR 2021}

\begin{document}

%%%%%%%%% TITLE
% \title{Identity Weighting Loss for Voice-Face Association Learning}
% \title{ Learning Voice-Face Association with Identity Alignment and Re-weighting}
\title{Seeking the Shape of Sound: An Adaptive Framework for \\Learning Voice-Face Association}
% Intermediate

% \and

\author{\parbox{16cm}
  {\centering
    {\large Peisong Wen$^{1,2}$ \ \ \ \ \ \ \ \ \  Qianqian Xu$^{1,}$\thanks{Corresponding authors} \ \ \ \ \ \ \ \ \  Yangbangyan Jiang$^{3,4}$ \\
    Zhiyong Yang$^{3,4}$ \ \ \ \ \ \ \ \ \ \ \   Yuan He$^{5}$ \ \ \ \ \ \ \ \ \ \ \ \ \ Qingming Huang$^{1,2,6,7,*}$ }\\
    {\normalsize
    $^1$ Key Lab of Intell. Info. Process., Inst. of Comput. Tech., CAS, Beijing, China\\
    $^2$ School of Computer Science and Tech., University of Chinese Academy of Sciences, Beijing, China \\
    $^3$ State Key Laboratory of Info. Security (SKLOIS), Inst. of Info. Engin., CAS, Beijing, China\\
    $^4$ School of Cyber Security, University of Chinese Academy of Sciences, Beijing, China \\
    $^5$ Alibaba Group, Beijing, China\\
    $^6$ BDKM, University of Chinese Academy of Sciences, Beijing, China\\
    $^7$ Peng Cheng Laboratory, Shenzhen, China\\
    % $^6$ Department of Mathematics, Hong Kong University of Science and Technology, Hong Kong\\
    }
    {\tt\small \{wenpeisong20z,xuqianqian\}@ict.ac.cn\quad\quad\{jiangyangbangyan,yangzhiyong\}@iie.ac.cn\quad\quad heyuan.hy@alibaba-inc.com \quad\quad qmhuang@ucas.ac.cn  
    }
  }
}

\maketitle

%%%%%%%%% ABSTRACT
\begin{abstract}
  Nowadays, we have witnessed the early progress on learning the association between voice and face automatically, which brings a new wave of studies to the computer vision community. However, most of the prior arts along this line 
  \textbf{(a)} merely adopt local information to perform modality alignment and \textbf{(b)} ignore the diversity of learning difficulty across different subjects.
  In this paper, we propose a novel framework to jointly address the above-mentioned issues. Targeting at \textbf{(a)}, we propose a two-level modality alignment loss where both global and local information are considered.  Compared with the existing methods, we introduce a global loss into the modality alignment process. The global component of the loss is driven by the identity classification. Theoretically, we show that minimizing the loss could maximize the distance between embeddings across different identities while minimizing the distance between embeddings belonging to the same identity, in a global sense (instead of a mini-batch). Targeting at \textbf{(b)}, we propose a dynamic reweighting scheme to better explore the hard but valuable identities while filtering out the unlearnable identities.
      % and realizes matching, verification and retrieval functions through embedding distances.
  % Specifically, we firstly introduce a two-level modality alignment for cross-modal embedding learning, which consists of implicit modality alignment and explicit modality alignment. In order to stabilize the learning process, we implicitly constrain the cross-modal distance by learning an intermediate modality. At the same time, we also improve the model with the explicit cross-modal alignment.
  % To handle the identity difference, we propose a strategy of adaptively re-weighting each identity in the loss functions, assigning larger weights to the hard identities and filtering out the abnormal identities.
  Experiments show that the proposed method outperforms the previous methods in multiple settings, including voice-face matching, verification and retrieval.
\end{abstract}

%%%%%%%%% BODY TEXT

% \documentclass[../cvpr.tex]{subfiles}
% \begin{document}
\section{Introduction}
\label{sec:intro}
% nature
Voice and face share various potential  characteristics, \eg, gender, ethnicity, age, which are helpful for identification and matching. Literatures \cite{smith2016matching, kamachi2003putting, mavica2013matching} show that humans can hear the voice of an unknown person and match the corresponding face with higher accuracy than chance, and vice versa. From the perspective of brain science, multimodal brain regions exist in the human brain, which process both voices and faces to form person identity representations \cite{tsantani2019brain}. Can machines learn such ability to recognize the face with the same identity only by hearing the voice, or recognize the voice from the face? In recent years, researchers have started to seek an answer to this interesting question \cite{nagrani2018seeing, wen2018disjoint}. The research of this technology is beneficial to many application scenarios, including criminal investigation, synthesis or retrieval of human faces from voices \cite{oh2019speech2face, wen2019face, bai2020speech, choi2019inference}, \etc. This task can be specialized as cross-modal matching, verification and retrieval problems. Different from the audio-visual speech recognition task \cite{zhang2020can}, the voice-face association problem is aim to find the identity relationship between face and voice, rather than the relationship between voice and facial action.

% previous work
In recent years, we have witnessed some progress of early studies along this line. As a representative example, SVHF \cite{nagrani2018seeing} regards the matching problem as a binary classification problem, and has achieved comparable performance with human baseline in both voice-to-face matching and face-to-voice matching. Benefit from the development of deep learning and the cross-modal retrieval technology, some recent work \cite{kim2018learning, wang2020learning, horiguchi2018face, xiong2019voice, nagrani2018learnable} has further verified the feasibility of this problem through deep metric learning. Wen \etal \cite{wen2018disjoint} boost the performance with multiple supervision.

\begin{figure}[t]
    \includegraphics[scale=0.43]{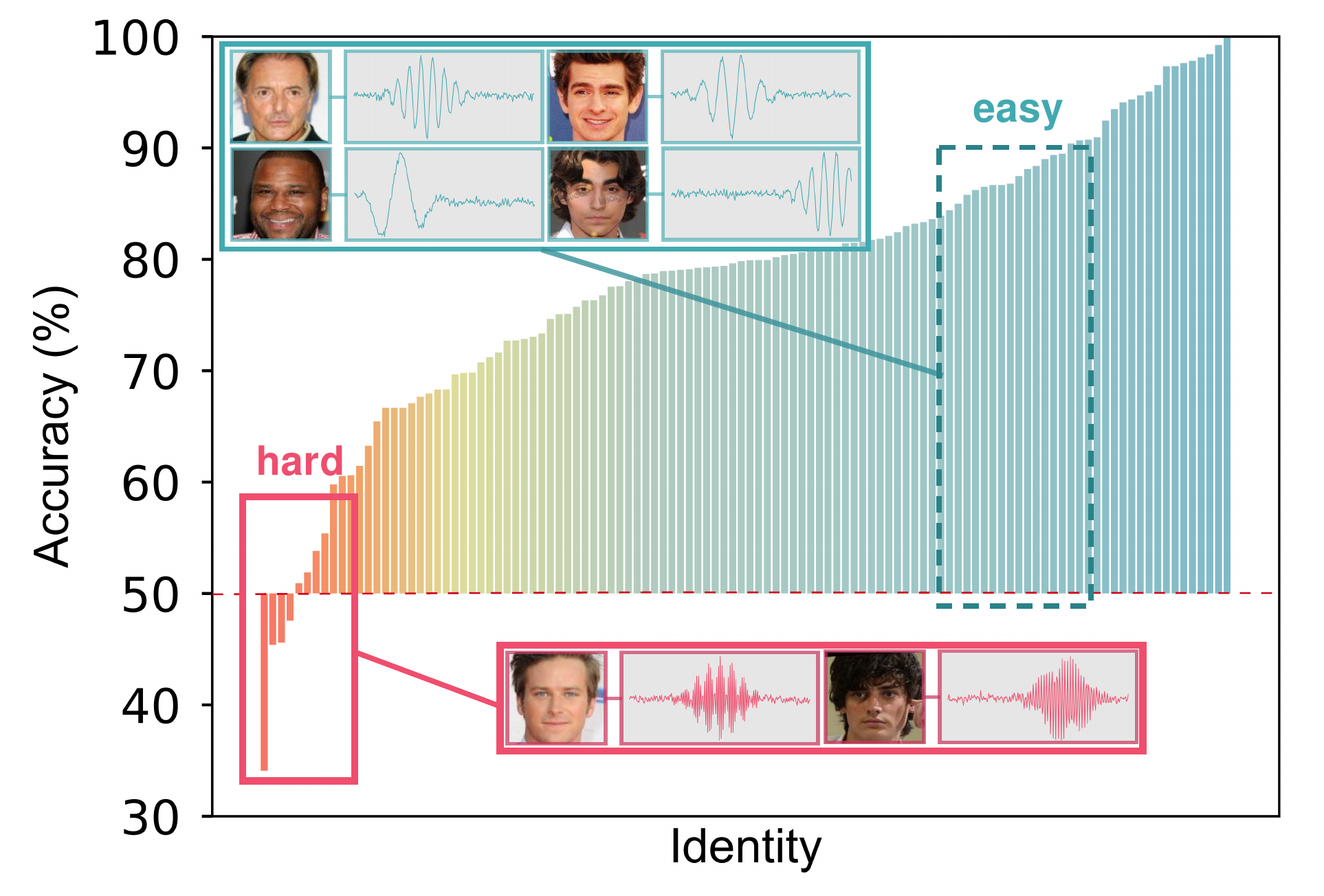}
    \centering
    \caption{Accuracy of different identities in the validation set under the 1:2 voice-to-face matching setting. There is a significant gap between identities, performances of some identities are even lower than chance (50\%).}
    \label{fig:acc_id}
\end{figure}

Despite previous methods have been able to reach the same level as untrained humans, there are still two problems in learning voice-face association. \textbf{(a)} The first problem is that contrastive loss functions used in previous work only use local information in a mini-batch, which leads to slow convergence. \textbf{(b)} The second one is that the diversity of difficulty across identities is ignored. Here the diversity of difficulty means that there are obvious differences in the difficulty of learning voice-face association among different subjects. To illustrate this problem, we train a model and test its accuracy on 1:2 voice-to-face matching for different subjects. The result is shown in \figref{fig:acc_id}, from which it can be noticed that the accuracy of identities is significantly different. This phenomenon coincides with what we find in reality. For example, not all male voices are low and rough, and there is an unignorable fraction of male voices that have their own characteristics. In \figref{fig:acc_id}, we show some  easy and hard identities in the obtained results. We observe that the accuracy distribution is relatively uniform, which validates our assumption that the learning difficulty is diverse. Moreover, there exists an unignorable fraction of identities suffering from a accuracy worse than random guess. This validates the existence of extremely hard identities.
%Taking the male identities for examples, there are some identities, featured with a significant high-frequency component of the voice, that sounds like a female.
Identities of this kind are hardly learnable. What is worse, they might even confuse the model and shift the correct decision boundary. In this paper, we name the hard but learnable identity as
\emph{hard identity}, and the extremely hard identity as \emph{personalized identity}. In this sense, a reasonable learning method should explore deeper into the hard identities while filtering out the personalized ones.

Based on the above consideration, in this paper, we propose an adaptive framework for the voice-face association learning. To overcome \textbf{(a)}, we introduce a two-level modality alignment, which consists of implicit and explicit modality alignment. The implicit alignment is implemented with an identity-classification-driven loss. With the theoretical analysis, we show that minimizing the implicit alignment loss could maximize the distance of embedding across modalities and identities and minimize the distance of embeddings across modalities but belong to the same identity. Moreover, the distance is measured from a global perspective instead of a local mini-batch. In this way, the implicit alignment introduces global information and identity semantics in the embeddings. Moreover, the explicit alignment, as a complementary component, aligns the two modalities in a mini-batch directly. For \textbf{(b)}, we propose an adaptive framework to handle the hard identities and personalized identities with dynamic identity weights. The hard identities obviously contribute to the bottleneck of the performance of the learning methods. We propose an adaptive weighting strategy to gradually increase the weights of the hard identities. This encourages the network to dive deeper into the associations between voice and face. Since personalized identities are extremely hard to learn, and their gradients are larger throughout the training phase, forcing the model to learn these samples will reduce the generalization of the model. Therefore, our proposed strategy adaptively assigns zero weights to personalized identities. 

In a nutshell, the main contributions of this work can be summarized as follows:
\begin{itemize}
    \item We propose explicit modality alignment and implicit modality alignment to effectively learn the voice-face association in a comprehensive manner.
    \item We propose an adaptive identity re-weighting framework to better explore  cross-modal associations from hard identities, and excluding personalized identities for generalization.
    \item Experiments under various settings are conducted to illustrate the effectiveness of the proposed framework.
\end{itemize}

% \documentclass[../cvpr.tex]{subfiles}
% \begin{document}
\section{Related Work}
\label{related_work}

\begin{figure*}[h]
    \centering
    \vspace{-1mm}
    \includegraphics[scale=0.82]{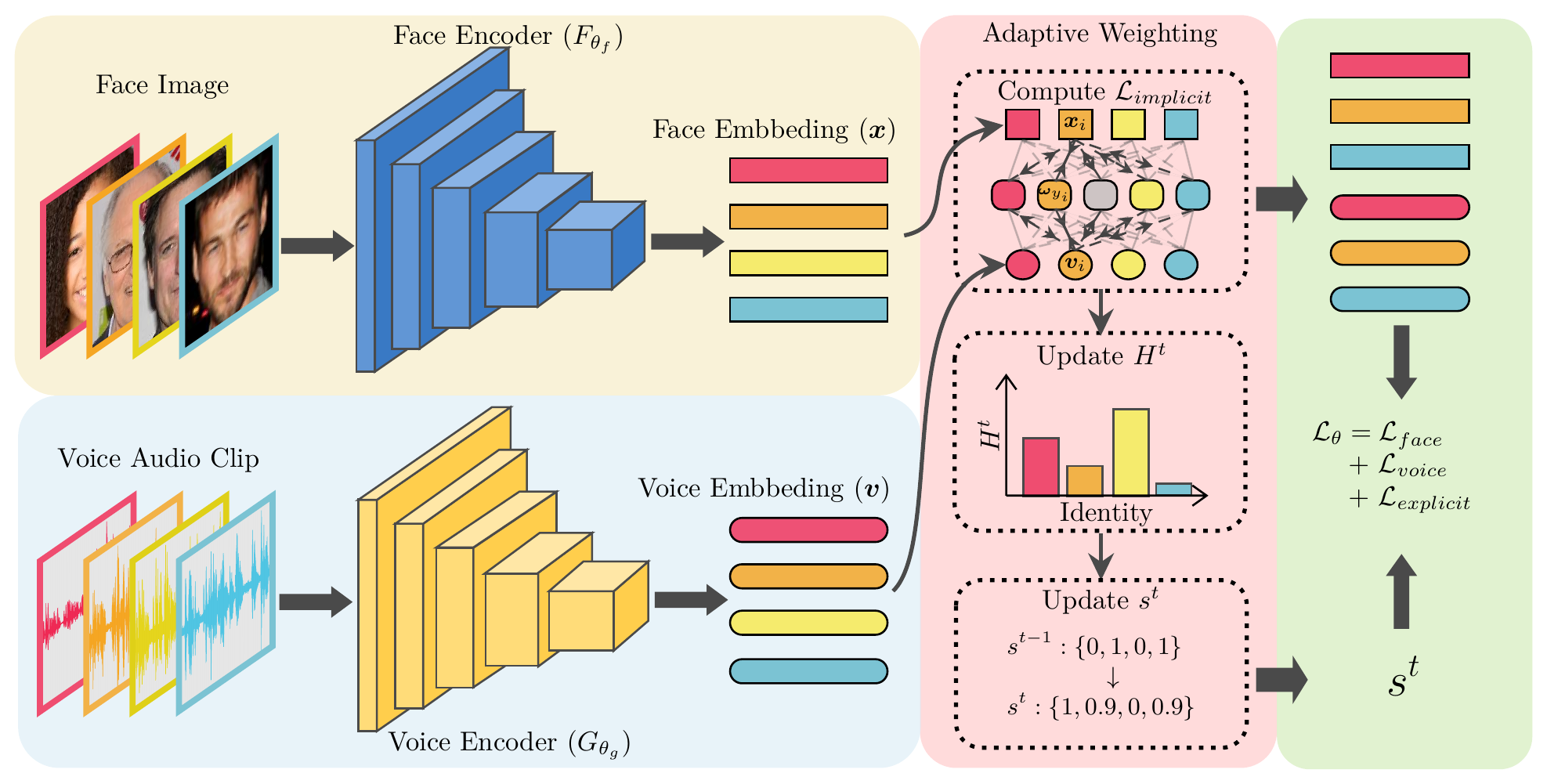}
    \caption{An overview of the proposed method. Face images and voice audio clips are fed into the face encoder network and the voice encoder network, respectively. The extracted embeddings are then assigned different weights according to the average loss of identities, and personalized samples are filtered out. Finally, the parameters of the network are updated with two-level modality alignment.}
    \label{fig:overview}
\end{figure*}

\subsection{Learning voice-face association}
In recent years, learning the voice-face association has aroused the interest of researchers. To the best of our knowledge, SVHF \cite{nagrani2018seeing} is the first work to propose a machine learning algorithm for the voice-face association learning, which focuses on the task of matching voices and faces. It poses the matching as a binary classification problem and simply uses the concatenation of the features as input of the classifier. Then the researchers turn to solve the problem from the cross-modal learning perspective, and include additional tasks such as cross-modal verification and retrieval. Inspired by existing cross-modal retrieval methods, some studies \cite{nagrani2018learnable,kim2018learning,xiong2019voice} adopt the metric learning technique, \eg, contrastive loss \cite{hadsell2006dimensionality} or triplet loss \cite{weinberger2006distance}, to help bridge the semantic gap between the two modalities. To further exploring the relationship between samples in a mini-batch, Horiguchi \etal utilize N-pair loss \cite{horiguchi2018face, sohn2016improved} and Wang \etal propose ranking loss \cite{wang2020learning} in the learning. Different from these approaches that are only supervised by the identity information, Wen \etal explicitly introduce more information like gender and nationality to supervise the learning in a multi-task manner \cite{wen2018disjoint}.

A shortcoming of most existing metric learning methods is that the losses they adopted only utilize the local information in a mini-batch. Yet in this work, we consider a two-level modality alignment where the global information in the whole dataset and the local information in the mini-batch are simultaneously employed.

\subsection{Sample re-weighting in deep learning}
In deep learning, different samples have different effects on model performance.
%In the process of model training, especially in the later stage of training, difficult samples can provide more effective gradients, but they may also bring negative effects.
To evaluate how samples affect models, Koh \& Liang \cite{koh2017understanding} use influence function to test the influence of sample weighting on loss without retraining. Based on the influence function, Wang \etal \cite{wang2018data} locate samples which are not helpful after the first round of training, and retrain the model from scratch without these samples. For the same purpose, Ren \etal \cite{ren2018learning} utilize meta learning method to assign weights based on the gradient direction of samples, and Fan \etal \cite{fan2017learning} explores deep reinforcement learning to automatically select samples in the training process. Sample re-weighting strategies are also applied in the object detection task \cite{shrivastava2016training,lin2017focal,li2019gradient}. One of the most well-known works is focal loss \cite{lin2017focal}, which alleviates class imbalance with a weighted loss function.

These re-weighting strategies only consider the instance-level difficulty, thus weight each sample separately. However, in the voice-face association learning task, the difficulty is identity-level. To address this issue, we propose a novel strategy that assigns identity-specific weights, where personalized identities are further excluded with zero weights adaptively. The advantages are two-fold: on one hand, it can reduce the number of weights and thus reduce the algorithm's complexity; on the other hand, the exclusion of personalized identities ensures the generalization of the network.

\section{Methodology}
% 2 pages
% For notational clarity
% possesses 

% \subsection{Problem definition}
% As is mentioned in \secref{sec:intro}, the voice-face association task can be embodied in the matching and retrieval between the two modalities. For description clarity, we only consider the 1:2 matching from voice to face in the following. Formally, given a dataset with $N_{train}$ samples: $\{(A_i, I_i, y_i)\}_{i=1}^{N_{train}}$, where $A_i$ and $I_i$ are voice audios and face images, respectively. Here $y_i \in \{1,2,\cdots,M\}$ are the identity labels, where $M$ is the total number of identities in the training set. Subscripts are omitted when there is no ambiguity. We have two functions $F_{\theta_f}$ and $G_{\theta_g}$ to map the raw data into $D$-dimensional embedding features: $x = F(I;\theta_f)$ and $v = G(A;\theta_g)$. The objective is to make as many positive samples as possible get higher score than negative samples under a similarity metrics:

Our goal is to learn generic vector representations to bridge the semantic gap across voices and faces such that a series of tasks ranging from face-voice/voice-face matching, verification to retrieval could be made available. To achieve this goal, we consider two factors. First of all, to narrow the gap across modalities, we expect to perform effective modality alignment. To this end, we introduce a loss-driven alignment mechanism  in \secref{sec:softmax} where the semantic gap is indirectly reduced by minimizing the identity classification loss function, which is called implicit modality alignment loss. %Theoretically, we find that such a procedure is approximately equivalent to maximizing the similarity of the embeddings across modalities.
Besides  indirect information, we also consider an explicit modality alignment loss in \secref{sec:cross} based on the contrastive learning framework. As for the second factor, we consider the diversity of learning difficulty mentioned in \secref{sec:intro}. It leads us to an adaptive learning framework with dynamic identity weights, which will be described in \secref{sec:noise}.

An overview of our framework is provided in \figref{fig:overview}, where 
we adopt the convolutional networks as our backbone.
% We implement the face encoder $F_{\theta_f}$ and the voice encoder $G_{\theta_g}$ with deep convolutional networks \cite{arandjelovic2016netvlad, zhong2018ghostvlad, Hu_2018_CVPR}.
After extracting embeddings, we perform an adaptive learning procedure via identity re-weighting. The loss function is the sum of the explicit and implicit modality alignment loss. When the identity weights are obtained, the final model is obtained by retraining the network based on the weights.
 
% \begin{figure}[h]
%     \includegraphics[scale=0.75]{tex/fig/fig_intermediate.pdf}
%     \caption{An illustration of intermediate modality (left) and explicit modality alignment (right). The former is implicit, but the intermediate modality is globally maintained, which is conducive to improving the training stability, while the explicit alignment is local, which can provide a powerful complementary.}
%     \label{fig:inter}
% \end{figure}

\subsection{Identity Recognition and Implicit Modality Alignment}
\label{sec:softmax}
First of all, we start with an implicit formulation of the modality alignment loss.
To make sure that the face/voice embeddings are consistent with the semantics of identity, we expect that the learned embeddings should lead to accurate identity recognition. This motivates us to minimize the softmax loss of the identity classification for both face and voice embeddings. Recalling Fig.\ref{fig:overview}, our backbone includes the deep encoders to leverage the face/voice embeddings and a linear classifier, i.e., the last FC layer of the architecture. We assume both voice and face embeddings share a common identity linear classifier. Its weight matrix is denoted as  $\bm{W} = [\bm{\omega}_1, \bm{\omega}_2, \cdots, \bm{\omega}_M] \in \mathbb{R}^{D\times M}$, where $M$ and $D$ represent the number of identities and feature dimensions respectively.
Given the training data $\mathcal{D} = \{(\vi, \xii, \yi)\}_{i=1}^{N}$, the identity classification loss is presented as follows:

\begin{eqnarray}
    \label{eq:l_face}
    &&\mathcal{L}_{face} = - \frac{1}{N} \sum\limits_{i=1}^N \log \frac{\exp(\wyi^T \xii)}{\sum_{j=1}^M \exp(\wj^T \xii)} \\
    \label{eq:l_voice}
    &&\mathcal{L}_{voice} = - \frac{1}{N} \sum\limits_{i=1}^N \log \frac{\exp(\wyi^T \vi)}{\sum_{j=1}^M \exp(\wj^T \vi)} \\
    \label{eq:l_exp}
    &&\mathcal{L}_{implicit} = \mathcal{L}_{face} + \mathcal{L}_{voice}
\end{eqnarray}
% where $\bm{W}_i$ refers to the $i$-th row of $\bm{W}$.
As an interesting fact, we can prove that adopting a common classifier could lead to an implicit modality alignment mechanism. This is shown in the following proposition.
\begin{prop}\label{prop:implicit}
Supposing that, for any $k \in \left\{1,2,\cdots,M \right\}$, the weight decay strategy ensures $\|\bm{\omega}_k\| \leq C$, we have a lower bound of $\mathcal{L}_{implicit}$ written as follows: 
\begin{align*}
 \mathcal{L}_{implicit}
 &\geq 2\log M - \frac{C}{MN}\sum_{j=1}^M D_j
                %  \overset{def}{=} {\mathcal{\tilde{L}}}_{implicit}
\end{align*}
where
\begin{align*}
    D_j &= \left\|(M - 1)\sum_{\yi = j}(\xii + \vi) - \sum_{\yi \neq j}(\xii + \vi)\right\|.
\end{align*}
\end{prop}

% \begin{rem}
    % We have the following remarks for Prop.\ref{prop:implicit}:
\noindent Prop. \ref{prop:implicit} shows the following properties of $\mathcal{L}_{implicit}$:
    \begin{itemize}
    \item According to the inequality in this proposition, minimizing $\mathcal{L}_{implicit}$ leads a smaller value of its lower bound $\log M - \frac{C}{MN}\sum_{j=1}^M D_j$, which equivalently leads to a larger value of $\frac{C}{MN}\sum_{j=1}^M D_j$.
    \item For a fixed $j$, $D_j$ is the overall distance between the face and voice embeddings belonging to $j$, and the embeddings that do not belong to $j$. Maximizing $\frac{C}{MN}\sum_{j=1}^M D_j$ eventually enforces all $\xii, \vi$ from the same class to be close to each other and enforces all $\xii, \vi$ from different classes to be far away from each other. In this sense,  minimizing $\mathcal{L}_{implicit}$ provides an \emph{implicit modality alignment} mechanism for our method.
    \item Compared with directly maximizing $\frac{C}{MN}\sum_{j=1}^M D_j$, minimizing $\mathcal{L}_{implicit}$ introduces $\bm{W}$ to the model which leverages  global information shared across all sample points. From the efficiency perspective, optimizing $D_k$ directly requires traversing the entire training set, while the implicit optimization can be accelerated by mini-batch training.
        % \item ${\color{org}(OP_2)}$ 
        % To integrate the discriminate information of the identities and the consistency across modalities.
    \end{itemize}

\subsection{Explicit modality alignment}
\label{sec:cross}

% Most of the studies for learning face-voice association are based on metric learning \cite{nagrani2018learnable,horiguchi2018face,kim2018learning,wang2020learning,xiong2019voice}. Although the loss functions based on paired data are unstable in training and difficult to converge when used alone, they are also better at avoiding falling into the local optimum. As discussed in \secref{sec:softmax}, learning cross-modal matching with softmax is actually implicit learning through the intermediate modality. Therefore, the parameters obtained are suboptimal solutions of the original problem.

% add fig here
In the previous subsection, we show that the identity recognition loss behaves like a global and implicit modality alignment loss. To obtain a comprehensive loss, we introduce a local and explicit modality alignment loss to include the complementary information of the implicit loss. 

Instead of using paired data directly, we implement the explicit modality alignment with N-pair loss \cite{sohn2016improved, sun2020circle}, which explores the local relationship among a mini-batch of instances. It is formulated as follows:
\begin{equation}
    \label{eq:l_cross}
    \begin{aligned}
    \mathcal{L}_{explicit} &= \frac{1}{N}\sum\limits_{i=1}^{N} \log(m + \frac{\sum_{\yj\neq \yi}\exp(\vi\xhatj)} {\exp(\vi\xhati)}) \\
    &+ \frac{1}{N}\sum\limits_{i=1}^{N} \log(m + \frac{\sum_{\yj\neq \yi}\exp(\xii\vhatj)} {\exp(\xii\vhati)})
    \end{aligned}
\end{equation}
where $m$ is a hyper parameter to control inter-class margins, and $\hat{\bm{x}} = \frac{\bm{x}}{\|\bm{x}\|}$, $\hat{\bm{v}} = \frac{\bm{v}}{\|\bm{v}\|}$. 
% This loss function is not symmetric for two modalities, but its symmetric form (exchange $x$ and $v$) also works well. It is enough to choose only one of them.

This loss function adds explicit constraints on the embeddings of the two modalities, as a powerful complementary to the implicit alignment. At first glance,  the ratio of positive samples and negative samples is $1:($N$-1)$, which might lead to the imbalance issue. Nonetheless, according to the following analysis, we can see that the loss terms will not be affected by the overwhelming ratio of the negative instances. In fact, $\mathcal{L}_{explicit}$ can be approximately written as 
\begin{equation}
    \begin{split}
        \mathcal{L}_{explicit}
        &\approx \frac{1}{N}\sum\limits_{i=1}^{N}[\max\limits_{\yj\neq \yi}\{\vi\xhatj\} - \vi\xhati + m-1]_+ \\
        & + \frac{1}{N}\sum\limits_{i=1}^{N}[\max\limits_{\yj\neq \yi}\{\xii\vhatj\} - \xii\vhati + m-1]_+
    \end{split}
\end{equation}
where $[x]_+$ indicates $\max(x,0)$. This shows that $\mathcal{L}_{explicit}$ behaves like a hinge loss to punish $\max\limits_{\yj\neq \yi}\{\vi\xhatj\} - \vi\xhati$. In this sense, only the one negative instance realizing the maximum is activated for each loss term. Hence, the imbalance issue could be naturally avoided.

% \begin{equation}
%     \mathcal{L}_{explicit} = \frac{1}{N}\sum\limits_{i=1}^{N} \log(m + \sum\limits_{y_j\neq y_i}\exp(v_i\hat{x_j}) / \exp(v_i\hat{x_i}))
% \end{equation}

\subsection{Adaptive identity re-weighting}
\label{sec:noise}

In this subsection, we propose an adaptive learning algorithm to deal with the diversity of the learning difficulties. The proposed method has three stages. During the first stage, we perform a warm-up training, where a pretrained model is learned without identity weights. In the second stage, the pretrained model is updated together with the identity weights. In the third stage, we train the final model based on the learned identity weights. A summary of all the details is shown in \AlgRef{alg:noise}.
% to handle hard identities and abnormal identities. We calculate the average loss of each identity, and give larger weights to those with larger average loss, while those with the largest average loss are treated as abnormal identities.

\begin{algorithm}
    \caption{Training with identity weights}
    \begin{algorithmic}[1]
        \REQUIRE {Training data $\mathcal{D}$, warm up iteration $T_{warm}$, update iteration $T_{update}$, max iteration $T_{max}$, batch size $N$, number of identities $M$, ratio of data retained $R_{keep}$.}
        \ENSURE {model parameters $\theta_f, \theta_g$.}
        \STATE{\color{org}\COMMENT{\texttt{First Stage}}}
        \STATE {\texttt{Train} $F_{\theta_f}, G_{\theta_v}$ \texttt{with} $\mathcal{D}$ \texttt{for} $T_{warm}$ \texttt{iterations.}}
        % \STATE {Calculate centers with \equref{eq:center} and initialize $s$.}
        % \STATE {$Q \leftarrow IsTopK (\frac{\bar{x}_i}{\|\bar{x}_i\|}\cdot\frac{\bar{v}_i}{\|\bar{v}_i\|}, 0.3)$.}
        % \STATE {Reinitialize $\theta_f, \theta_g$.}
        % \FOR{$t=1$ to $T_{max}$}
        % \STATE{}
        \STATE{}{\color{org}\COMMENT{\texttt{Second Stage}}}
        % \WHILE {$R_{keep} > 0$}
        \WHILE {$\sum\limits_{i=0}^M I[s_i^{t-1} > 0] < R_{keep}\times M$}
        \STATE {$\{(A_i, I_i, y_i)\}_{i=1}^N \leftarrow $\texttt{SampleMiniBatch}($\mathcal{D}, N$)}
        \STATE {$\xii, \vi \leftarrow F_{\theta_f}(I_i), G_{\theta_g}(A_i)$.}
        \STATE {\texttt{Calculate} $\mathcal{L}_{implicit}$ \texttt{with} \equref{eq:l_face} \texttt{and} (\ref{eq:l_voice}).}
        \STATE {\texttt{Update} $H^{t}$ \texttt{with} \equref{eq:mean_loss}.}
        % \STATE {Update centers with \equref{eq:l_center}.}
        \IF {$t \% T_{update} = 0$}
            \STATE {\texttt{Update} $s^{t}$ \texttt{with} \equref{eq:s_update}.}
            % \STATE { $R_{keep}\leftarrow R_{keep} - k/M$}
        \ELSE
            \STATE {$s^{t} \leftarrow s^{t-1}$.}
        \ENDIF
        \STATE {\texttt{Update} $\theta_f,\theta_g$ \texttt{with} \equref{eq:l_theta}.}
        \ENDWHILE
        %  \STATE{}        
        \STATE{}{\color{org}\COMMENT{\texttt{Third Stage}}}
        \STATE {\texttt{Reinitialize} $\theta_f, \theta_g$.}
        \FOR{$t=1$ to $T_{max}$}
        \STATE {\texttt{Update} $\theta_f,\theta_g$ \texttt{with} \equref{eq:l_theta}.}
        \ENDFOR
    \end{algorithmic}
    \label{alg:noise}
\end{algorithm}

The details for the first and third stage are obvious. We only elaborate on the second stage next.

During this training phase, we update the weights in an iterative manner. In the $t$-th iteration, we first sample a mini-batch (\texttt{Line 5} in \AlgRef{alg:noise}) and perform a standard inference to obtain the embeddings $\xii, \vi$ for all the instances in the mini-batch (\texttt{Line 6} in \AlgRef{alg:noise}). Next, we evaluate the hardness of each involved identity in the current mini-batch. To ensure the robustness of the learned weights, we only employ the more stable implicit loss (\texttt{Line 7} in \AlgRef{alg:noise}) $\mathcal{L}_{implicit}$ to measure the hardness. To integrate the hardness information from previous iterations and the current iteration, we perform a moving average strategy to calculate the hardness (\texttt{Line 8} in \AlgRef{alg:noise}). Above all, the hardness for the $i$-th identity, \ie, $H^{t}_{i}$,  could be written as:
\begin{equation}
    \label{eq:mean_loss}
    H^{t}_{i} = \beta \cdot H^{t-1}_{i} + (1 - \beta) \cdot \mathcal{L}_{implicit}
\end{equation}
where $\beta \in (0,1)$ is a hyper parameter to control the importance of the current iteration.

Now, we assign a weight for each identity to gradually add the hard identities to the training set (\texttt{Line 11} in \AlgRef{alg:noise}). Specifically, in the $t$ iteration of the training process, the weight of the samples belonging to the $i$-th identity in the loss is denoted as $s^{t}_i$. As for the initialization, $s^{0}_i$ is set to $1$ for the bottom 30\% $H^{0}_{i}$, and is set to $0$ otherwise. To maintain
\noindent stability, the weight update is triggered after every $T_{update}$ iterations (\texttt{Line 9} in \AlgRef{alg:noise}). If the update is triggered, 
$s^{t}_i$ will be updated as follows:

\begin{equation}
    \label{eq:s_update}
    s^{t}_i =
    \begin{cases}
        1, \emph{for all bottom} ~k~ H^{t}_{i}~\emph{ with} ~s^{t-1}_i = 0. 
     \\
        \alpha \cdot s_i^{t-1}, \emph{otherwise.}
    \end{cases}
\end{equation}
where $\alpha \in (0,1)$ and $k$ (a positive integer) are hyper parameters. Next, we explain how this equation works. First, we will pick out all the identities with the smallest $k$ $H^{t}_i$ and set their $s^t_i$ to $1$. For the rest of the identities, we simply employ a weight decay strategy by setting $s^t_i  =\alpha\cdot s^{t-1}_i$.
Meanwhile, the identities with the largest $H^t_i$ are considered as personalized identities, which should be dropped out via a zero weight. To this end, we update the identity weights sequentially. We will terminate the second stage of training immediately after $R_{keep}$ percent of identities are assigned with non-zero weights (\texttt{Line 4} in \AlgRef{alg:noise}).

At the end of each iteration (if not terminated), the network parameters $\theta_f, \theta_g$ are updated with weighted loss functions (\texttt{Line 14} in \AlgRef{alg:noise}):
\begin{eqnarray}
    \label{eq:l_face_denoise}
    &&\mathcal{L}_{face} = - \sum\limits_{i=1}^N \hat{s}^{t}_{\yi} \log \frac{\exp(\wyi \xii)}{\sum_{j=1}^M \exp(\wj \xii)} \\
    \label{eq:l_voice_denoise}
    &&\mathcal{L}_{voice} = - \sum\limits_{i=1}^N \hat{s}^{t}_{\yi} \log \frac{\exp(\wyi\vi)}{\sum_{j=1}^M \exp(\wj \vi)}
    \label{eq:l_cross_denoise}
    % \notag
\end{eqnarray}
\begin{eqnarray}
    &&\begin{aligned}
        \mathcal{L}_{explicit} = \sum\limits_{i=1}^{N} \hat{s}^{t}_{\yi} \log(m + \frac{\sum\limits_{\yj\neq \yi}\exp(\vi\xhatj)} {\exp(\vi\xhati)}) \\
        + \sum\limits_{i=1}^{N} \hat{s}^{t}_{\yi} \log(m + \frac{\sum\limits_{\yj\neq \yi}\exp(\xii\vhatj)} {\exp(\xii\vhati)})
    \end{aligned} \\
    \label{eq:l_theta}
    &&\mathcal{L_{\theta}} = \mathcal{L}_{face} + \mathcal{L}_{voice} + \mathcal{L}_{explicit}
\end{eqnarray}
where $\hat{s}^t_i = s^t_i/\sum\limits_{j=1}^Ns^{t}_j$.

% To this end, we propose a training pipeline to exclude noisy identities, and assign higher weights for hard identities, as shown in \AlgRef{alg:noise}. Relevant researches show that noisy samples are more difficult to converge than clean samples. 

% We firstly train the model with the entire training set for a few iterations, and then calculate average loss for each identity in the training set as follows:
% \begin{equation}
%     \label{eq:mean_loss}
%     L_{mean} = \beta L_{mean} + (1 - \beta)\mathcal{L}_{implicit}^{(t)}
% \end{equation}
% where $\beta$ is a hyper parameter to control update rate.

\section{Experiments}
\subsection{Datasets}
Following previous work \cite{wen2018disjoint, wang2020learning}, we evaluate the proposed method on a constructed dataset based on VoxCeleb \cite{nagrani2017voxceleb} and VGGFace \cite{parkhi2015deep} datasets. VoxCeleb is an audio-visual dataset of short human speech videos, which provides audios for our experiments. Meanwhile, VGGFace is a human face dataset of 2,622 identities. Since in our settings, it is not necessary to capture voices and faces from the same video, we use still faces from VGGFace instead of extracted faces from VoxCeleb. The intersection of VoxCeleb and VGGFace contains 1,225 identities after filtering low quality data. Then we split the data into train/validation/test sets without identity overlapping, and generate the queries for validation and test according to our evaluation protocols (See \secref{subsec:implement}). The statistics of the resulted dataset are reported in \tabref{tab:dataset}.
%In the 1:$N$ matching setting, a gallery consists of a randomly selected positive sample and $N-1$ negative samples from other identity. For each instance in the test set, we randomly generate 3 galleries. The verification testing list is generated based on the 1:2 matching, where a randomly selected sample from a gallery is used to query. We use all instances for retrieval.

\begin{table}[h]
	\caption{Details on the datasets.}
    \centering
    \setlength{\tabcolsep}{5pt}
    \begin{tabular}{c|cccc}
    \toprule
                & train     & validation & test   & total \\
    \midrule
    audio clips & 113,322   & 14,182     & 21,850 & 149,354 \\
    face images & 104,724   & 12,260     & 20,076 & 137,060 \\
    identities  & 924       & 112        & 189    & 1,225 \\
    queries (V-F) & --—  & 42,546     & 65,550 & 108,096 \\
    queries (F-V) & --—  & 36,780     & 60,228 & 97,008 \\
    \bottomrule
    \end{tabular}
    \label{tab:dataset}
\end{table}

\begin{table*}[t]
	\caption{Results (\%) on 1:2 matching, verification and retrieval. V-F: from voice to face; F-V: from face to voice; U: unrestricted; G: query restricted by gender. The best results of our models and competitors are highlighted in \first{soft red} and \second{soft blue}, respectively.}
    \centering
    \setlength{\tabcolsep}{5pt}
    \begin{tabular}{c|cccc|cccc|cc}
    \toprule
    \multirow{2}{*}{Methods}
    & \multicolumn{4}{c}{1:2 Matching (ACC)}
    & \multicolumn{4}{|c}{Verification (AUC)}
    & \multicolumn{2}{|c}{Retrieval (mAP)} \\
    
    \cline{2-11} & V-F (U) & F-V (U) & V-F (G) & F-V (G)
                & V-F (U) & F-V (U) & V-F (G) & F-V (G)
                & V-F & F-V \\
    \midrule
    SVHF \cite{nagrani2018seeing} & 81.0 & 79.5 & 63.9 & 63.4 & --— &--—&--—&--—&--—&--—\\
    DIMNet \cite{wen2018disjoint} & 81.3 & 81.9 & 70.6 & 69.9 & 81.0 & 81.2 & \second{70.4} & 69.3 & 4.3 & \second{3.8} \\
    Wang's \cite{wang2020learning} & \second{83.4} & \second{84.2} & \second{71.7} & \second{71.1} & \second{82.6} & \second{82.9} & 70.3 & \second{70.1} & \second{4.4} & 3.4\\
    \midrule
    Ours (focal \cite{lin2017focal}) & 85.3 & 85.0 & 75.6 & 74.5 & 85.6 & 85.4 & 76.0 & 75.1 & 6.5 & 6.2 \\
    Ours (-E -W) & 84.6 & 84.7 & 72.8 & 71.3 & 84.8 & 85.0 & 72.4 & 71.4 & 4.6 & 5.1 \\
    Ours (-E) & 84.3 & 84.3 & 75.0 & 74.4 & 84.6 & 84.6 & 75.3 & 74.8 & 5.1 & 4.9 \\
    Ours (-I -W) & 85.1 & 85.4 & 75.2 & 74.3 & 85.7  & 85.7 & 75.8 & 75.2 & 4.8 & 5.1 \\ 
    Ours (-W) & 85.5 & 85.2 & 76.3 & 74.7 & 85.8 & 85.3 & 76.7 & 75.1 & \first{6.5} & \first{6.2} \\
    Ours & \first{87.2} & \first{86.5} & \first{77.7} & \first{75.3} & \first{87.2} & \first{87.0} & \first{77.5} & \first{76.1} & 5.5 & 5.8 \\
    \bottomrule
    \end{tabular}
    \label{tab:res_m_v_r}
\end{table*}

\subsection{Implementation details} \label{subsec:implement}
\noindent \textbf{Network architecture.}
The face feature extractor is implemented with SE-ResNet-50 \cite{hu2018squeeze}, which is a powerful backbone on multiple tasks of computer vision. The input is a face image of size $112\times112\times3$, which is normalized to $[-1, 1]$ by subtracting $127.5$ and dividing $127.5$, and the output is a 128-dimensional face embedding. The voice feature extractor is implemented with Thin-ResNet-34 \cite{arandjelovic2016netvlad, zhong2018ghostvlad}, which inputs a spectrogram of voice and outputs a 128-dimensional voice embedding. Spectrograms have 257 channels (256 frequency components and 1 DC component), which are generated with a hamming sliding window of width 25ms and of hop 10ms. The two networks are pre-trained with the face recognition task on MS-1M \cite{guo2016ms}, and the audio speaker recognition on VoxCeleb2 \cite{chung2018voxceleb2}, respectively.
% Then the embeddings are fed into the shared classifier, which is implemented as a single full-connected layer here, to obtain the final prediction.

\noindent \textbf{Training strategy.}
The training process is split into three stages: warm-up training, identity weight learning and retraining with fixed identity weights, as mentioned in \secref{sec:noise}. In order to balance the number of samples with different identities, we sample several identities in an iteration, and then randomly sample one face image and one voice audio clip for each identity. For data pre-processing, we apply data augmentation including random rotation (from $-15^{\circ}$ to $15^{\circ}$), random cropping ($112\times 112$) for images, and random cropping in time axis (from 2.5 seconds to 5.0 seconds) for audios. In our model, the hyperparameters are set as follows: $m = 3.4$ (\equref{eq:l_cross_denoise}), $\beta = 0.9$ (\equref{eq:mean_loss}), $\alpha = 0.99, k = 22$ (\equref{eq:s_update}), $T_{warm} = 500, T_{update} = 100, R_{keep}=0.9$ (\AlgRef{alg:noise}). We adopt the stochastic gradient descent (SGD) optimizer, where batch size and momentum are set to $64$ and $0.9$, respectively. The learning rate is initialized as $10^{-2}$, and decays by $0.1$ in the 2k and 3k iterations. The max iteration $T_{max}$ is 10k, and the best model on the validation set is preserved for evaluation.

\begin{figure*}[t]
    \vspace{2mm}
    \subfigure{
        \begin{minipage}{1.0\textwidth}
            \centering
            \centerline{\includegraphics[scale=0.20]{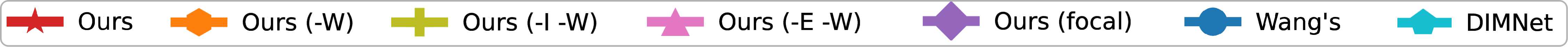}}
        \end{minipage}
	}

    \vspace{-3.5mm}
	\subfigure{
		\begin{minipage}[b]{0.32\textwidth}
			\includegraphics[scale=0.37]{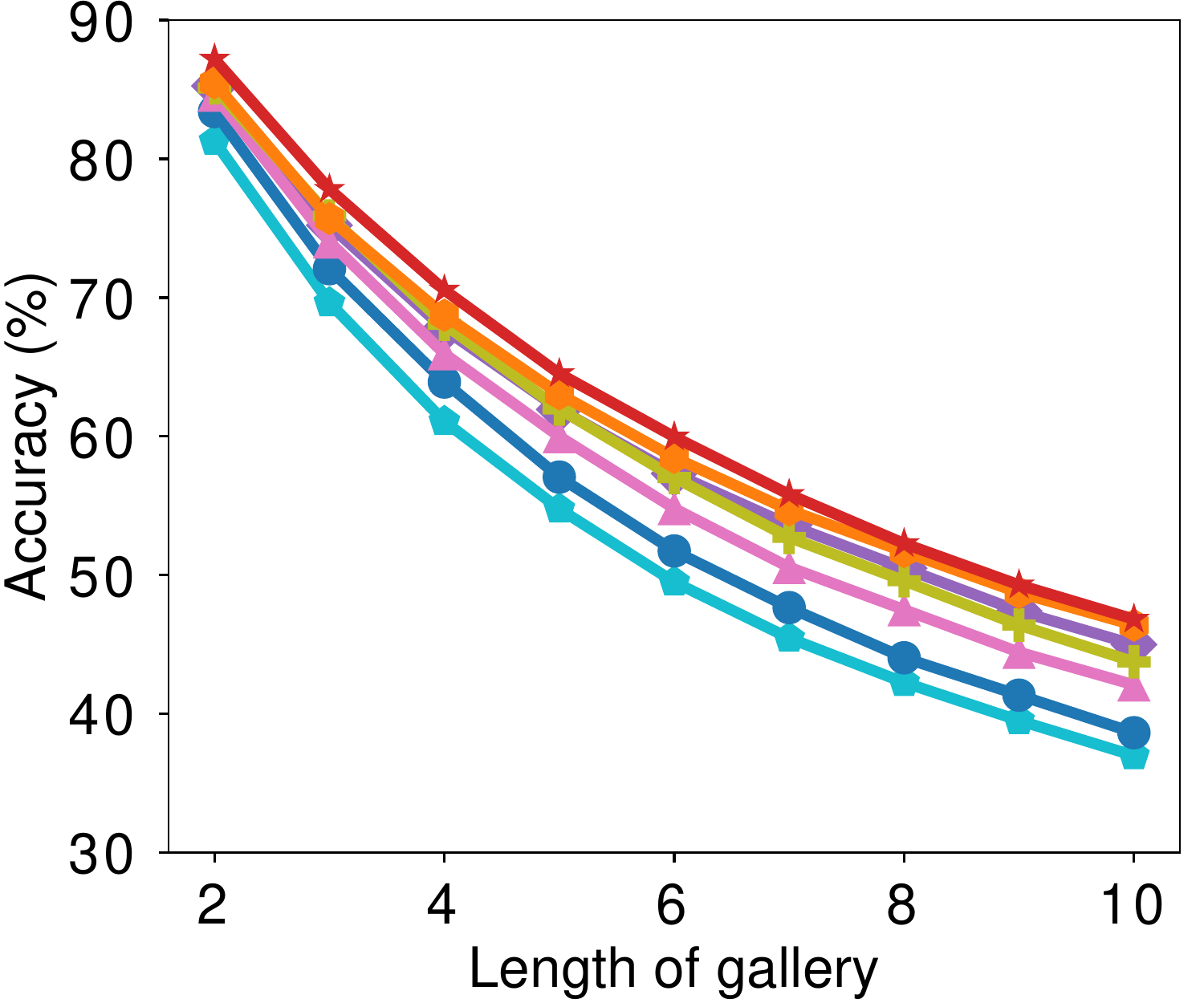}
            \centerline{(a) V-F (U)}
        \end{minipage}
		\label{fig:match_v2f}
	}
    % \hspace{.1in}
	\subfigure{
		\begin{minipage}[b]{0.32\textwidth}
            \includegraphics[scale=0.37]{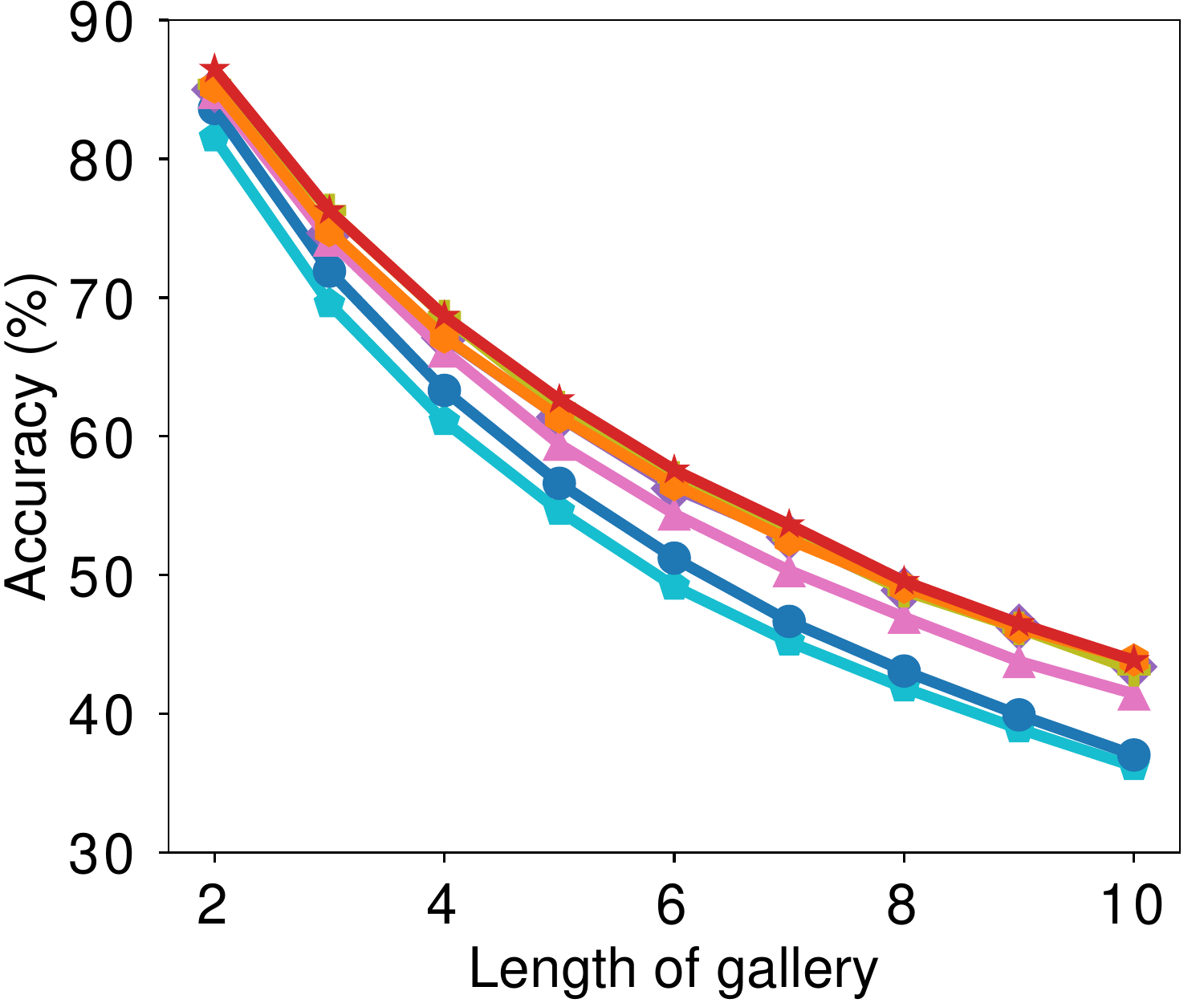}
            \centerline{(b) F-V (U)}
		\end{minipage}
		\label{fig:match_f2v}
	}
    % \hspace{.1in}
    % 
	\subfigure{
		\begin{minipage}[b]{0.32\textwidth}
			\includegraphics[scale=0.37]{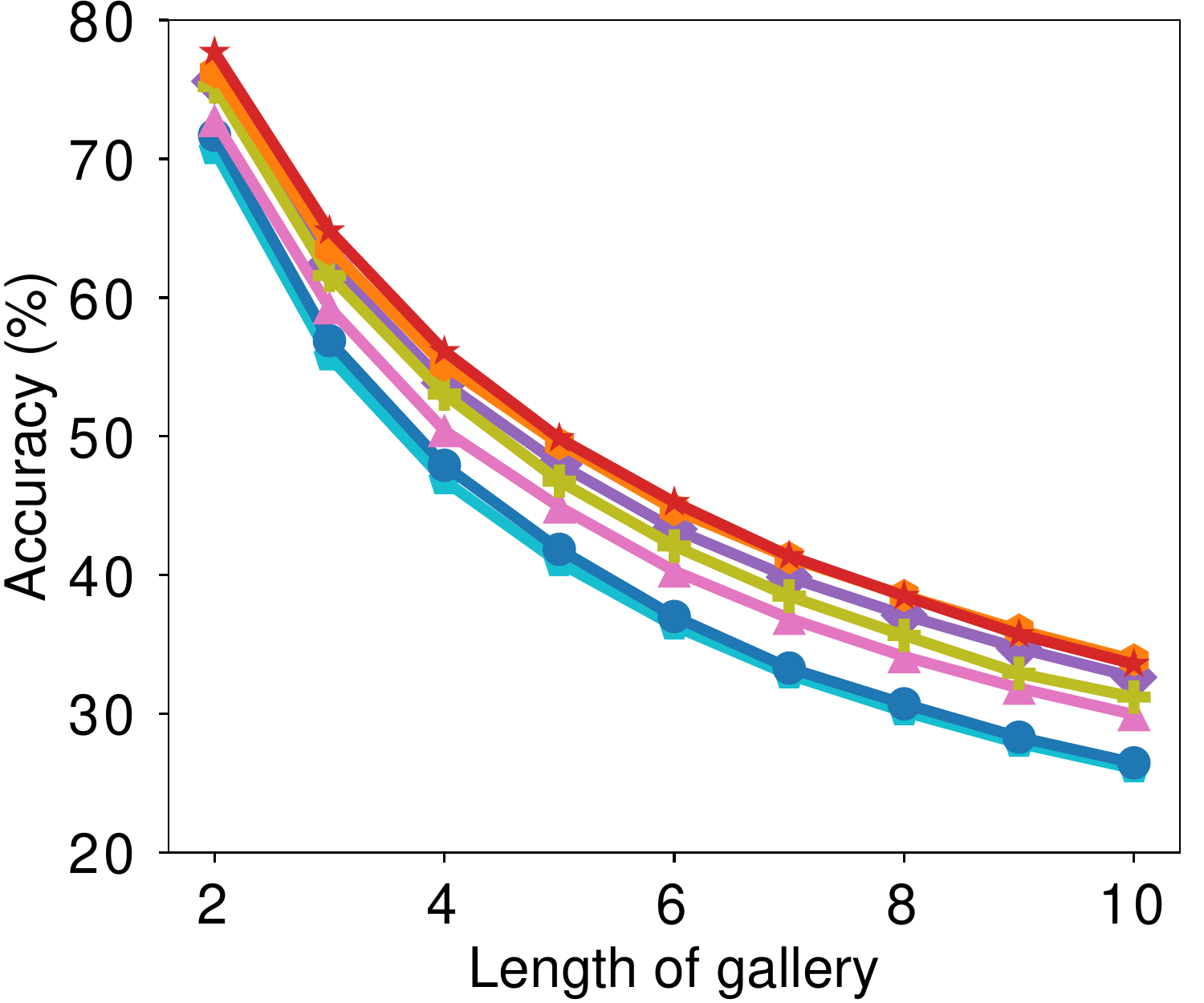}
            \centerline{(c) V-F (G)}
		\end{minipage}
		\label{fig:match_g_v2f}
    }
	% \hspace{.1in}
	% \subfigure[F-V (G)]{
	%         \begin{minipage}[b]{0.32\textwidth}
	%     \includegraphics[scale=0.37]{tex/fig/match_g_f2v.eps}
	%         \end{minipage}
	%     \label{fig:match_g_f2v}
    % }
	\caption{Quantitative results on 1:$N$ matching task. Best viewed in color.}
	\label{fig:match_n}
\end{figure*}

\noindent \textbf{Evaluation protocol.}
To show the overall capacity of the proposed model in voice-face association learning task, we conduct the evaluation under the following four settings:
\begin{enumerate}[(a)]
    \item \textbf{1:2 matching}. Given an instance from one modality as the probe, and two candidates from the other modality (including one from the same identity as the probe) as the gallery, the task is to find out which candidate matches the probe. The performance is measured with accuracy (ACC).
    \item \textbf{1:$\mathbf{N}$ matching}. This task is basically the same as 1:2 matching, except that the length of gallery $N$ ranges from $2$ to $10$ in our experiments. We report ACC on each $N$.
    \item \textbf{Verification}. Given two instances from different modalities, the task is to judge whether they belong to the same person. The performance is measured with Area Under the ROC curve (AUC).
    \item \textbf{Retrieval}. This task is extended from the 1:$N$ matching task, where the gallery contains one or more candidates that match the probe. The model is asked to rank the gallery samples so that the candidates matching the probe are ranked at the top. We report the performance with mean average precision (mAP).
\end{enumerate}
For each task, we report the metrics for two types of queries, from voice to face (\textbf{V-F}) and from face to voice (\textbf{F-V}). On matching and verification tasks, the queries are further divided into two subtypes: gallery samples have the same gender as the probe sample (\textbf{G}), or have unrestricted genders (\textbf{U}).

\noindent \textbf{Competitors.} We compare our proposed method with three models: SVHF \cite{nagrani2018seeing}, DIMNet \cite{wen2018disjoint} and Wang's model \cite{wang2020learning}. Note that SVHF takes dynamic faces as input, so its dataset is not exactly the same as ours.

\noindent \textbf{Ablated variants.} In order to quantify two main contributions of this paper: two-level modality alignment and identity re-weighting, we also implement four ablated variants: (1) the model without explicit alignment and re-weighting (denoted by \textbf{-E -W}), (2)the model without implicit alignment and re-weighting (denoted by \textbf{-I -W}) (3) the model without re-weighting (denoted by \textbf{-W}), and (4) a model trained with focal loss \cite{lin2017focal} instead of our re-weighting strategy (denoted by \textbf{focal}).

\subsection{Results and comparison}
\noindent \textbf{Quantitative results.} The results on the 1:2 matching, verification and retrieval are recorded in \tabref{tab:res_m_v_r}. We see that our model significantly outperforms the competitors over all three tasks, with an average improvement of about 2\%-7\%. Furthermore, comparing the performance on gender restricted and unrestricted queries, our method has a larger improvement in the case of gender restriction. These validate that our model could discover deeper associations between face and voice. On the other hand, the results of 1:$N$ matching are shown in \figref{fig:match_n}, which further verifies the advantage of our method. In this task, the accuracy decreases with the increase of N, but our method consistently has a higher performance and has less decrease than the competitors, which shows that the proposed framework is relatively more robust. 

Moreover, we could make the following observations: (1) The explicit modality alignment can significantly improve performance, especially in the gender-restricted groups, which are more challenging. In the 1:2 matching task and the verification task, identity re-weighting brings an increase of about 1\%, and achieves the state-of-the-art performance. However, training with focal loss doesn't bring significant improvement. (2) For all the tasks, results on gender-unrestricted queries are obviously better than gender-restricted ones. This shows that facial and voice features are strongly related to gender, which is consistent with people's experience. Besides, even if the gender is restricted, machines can still learn the association between face and voice. (3) Despite that our identity re-weighting strategy brings improvements on the matching and verification tasks, the full model does not perform as well as its ablated counterpart on the retrieval task.
% Similar phenomenon can also be observed on 1:$N$ matching, especially in \figref{fig:match_n}{\color{red}(c)}.
One possible reason is that the re-weighting strategy improves generalization at the expense of precision on the tail of gallery.
%This is one of the shortcomings of our model at present.
How to improve the performance of this situation more effectively is worthy of further study.

% while in retrieval, every query needs to face all data, including those from abnormal identities, which makes it more likely to be disturbed.

\begin{figure}[t]
    \centering
    \vspace{-1mm}
	\subfigure[Embeddings on training set.]{
		\includegraphics[scale=0.24]{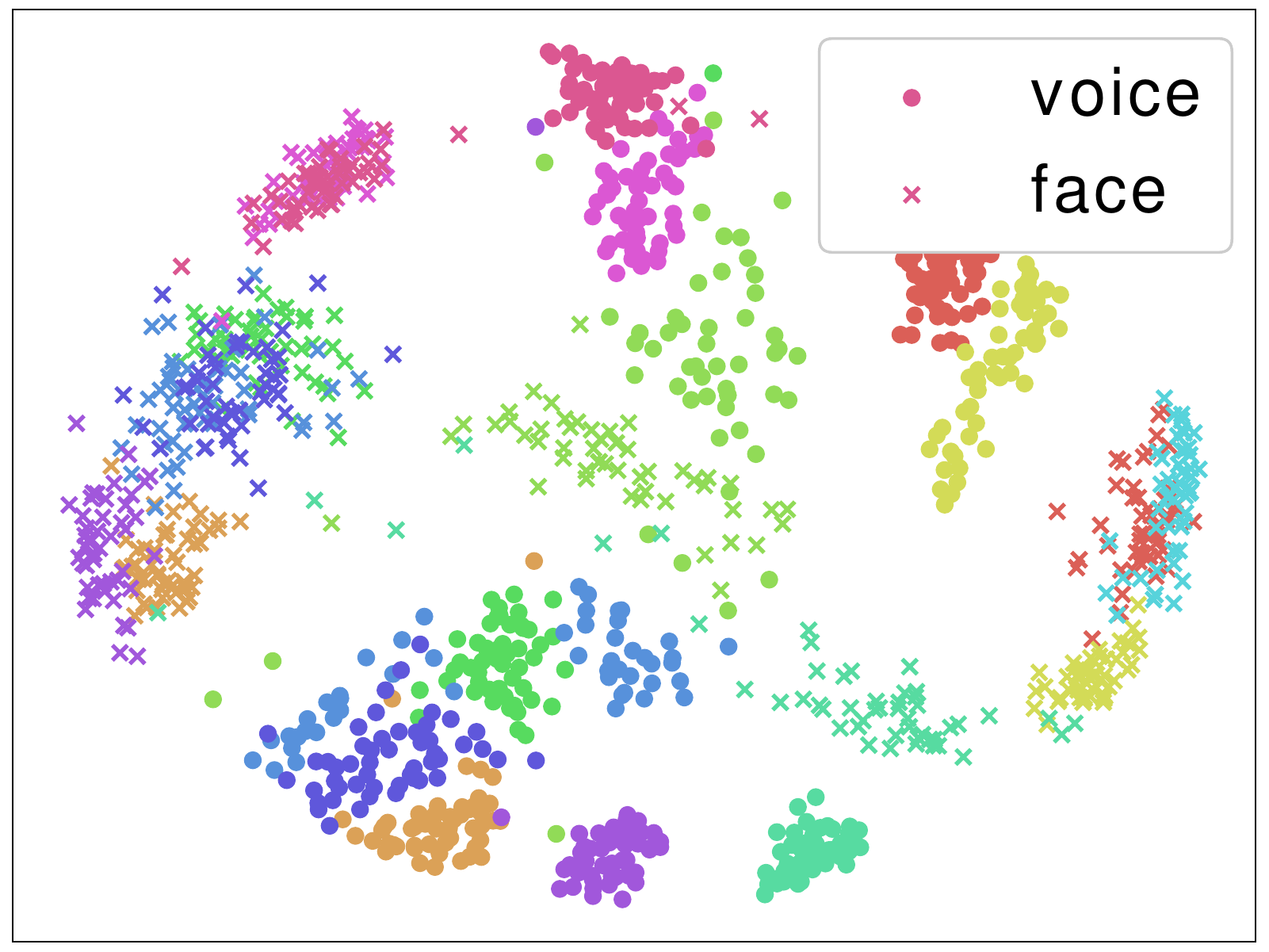}
		\label{fig:emb_train}
	}
	\subfigure[Embeddings on test set.]{
		\includegraphics[scale=0.24]{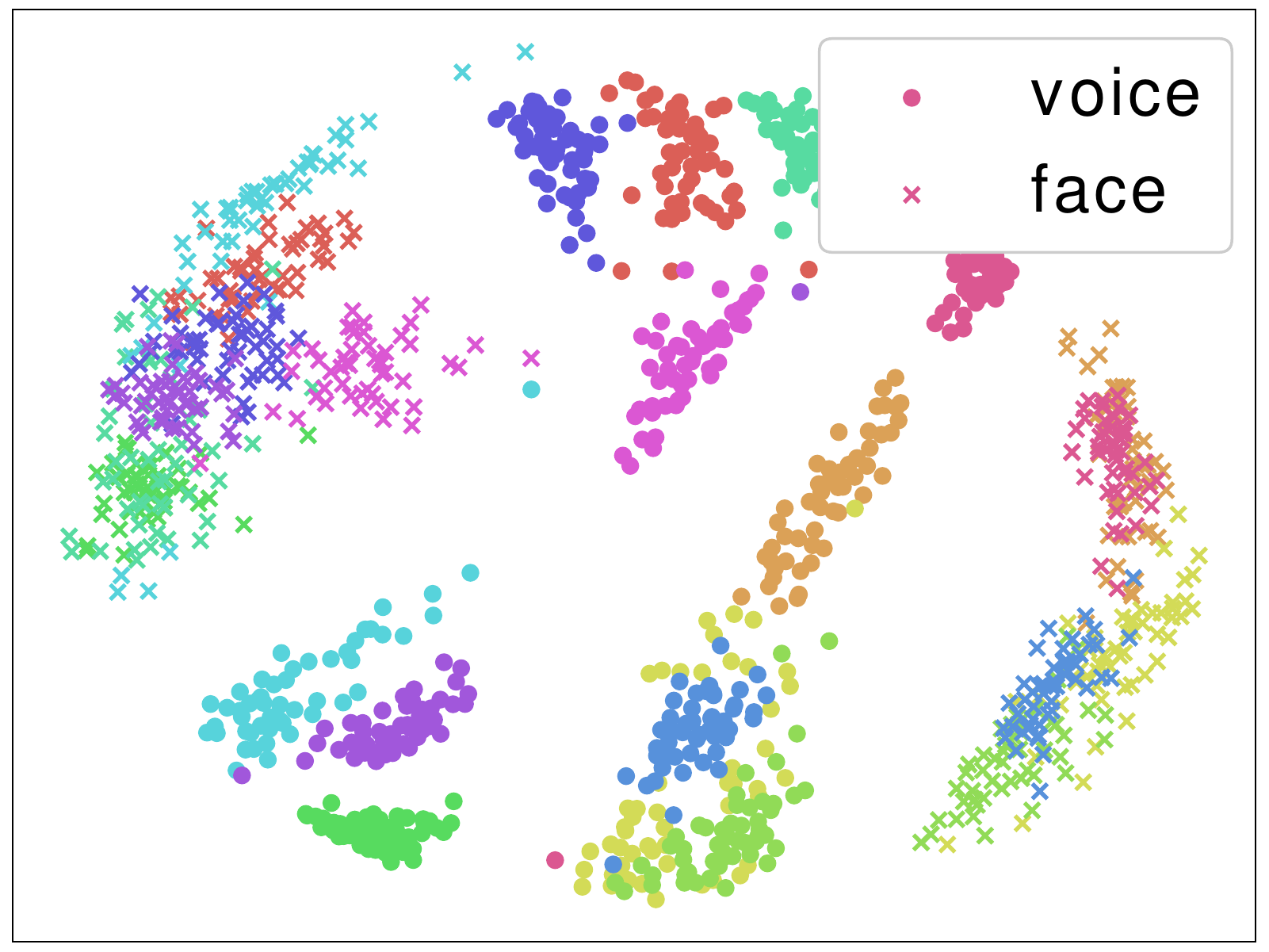}
		\label{fig:emb_test}
	}
	\caption{Visualization of learned embeddings. The same color points belong to the same person, and different shapes represent different modalities. Dimension reduction is implemented by multi-dimensional scaling (MDS) \cite{wickelmaier2003introduction}. Best viewed in color.}
	\label{fig:emb}
\end{figure}

\noindent\textbf{Qualitative results.} A visualization of the face embeddings and the voice embeddings extracted with our model is provided in \figref{fig:emb}. It could be observed that embeddings from the same identities are close and discriminative in most cases, which shows the effectiveness of learned features intuitively.

%%% 1:N matching + fig
%%% write more here

\begin{figure}[t]
    \centering
    \includegraphics[scale=0.9]{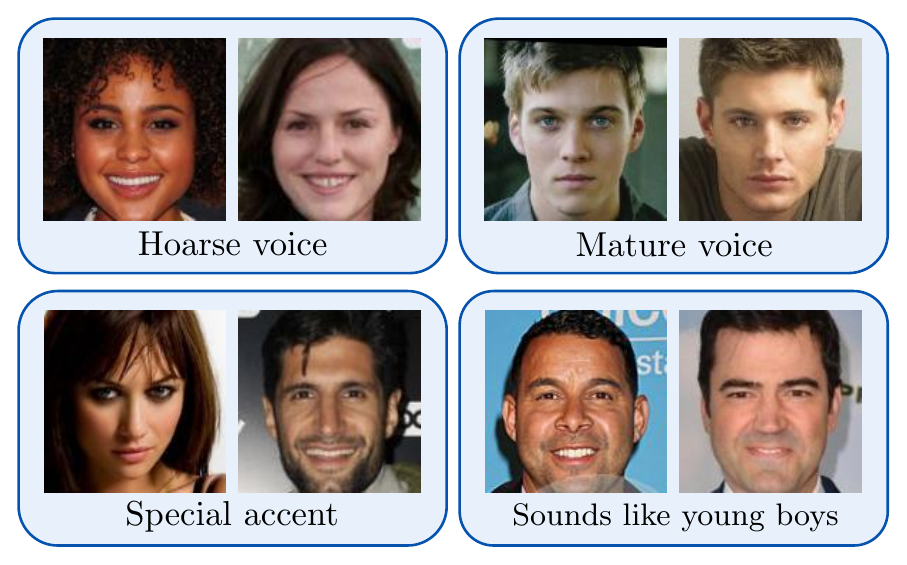}
    \caption{Some examples of personalized samples. We show four cases of voice and face inconsistency as examples.}
    \label{fig:noise_sample}
\end{figure}

\subsection{Ablation study for training strategy}
\noindent \textbf{Effect of identity weights.}
In our framework, the hard identities are filtered out by manually setting the ratio of data retained $R_{keep}$. As is shown in \tabref{tab:keep_ratio}, we evaluate the 1:2 matching accuracy of models learned with different $R_{keep}$, ranging from 0.6 to 1.0. Three conclusions can be drawn: (1) Weighting the difficult samples can bring about 1.2\% to 1.6\% performance improvement, even if extremely hard identities are not excluded. (2) After dropping extremely hard identities, the performance is further boosted, verifying the effectiveness of eliminating personalized identities. (3) After a certain point, the performance of the model decreases with the increase of the retaining ratio. This is due to the abandonment of some data might lower the data diversity. Some excluded samples are shown in \figref{fig:noise_sample}. Intuitively speaking, these samples are usually special in voice. See {\it Supplementary Materials} for more analysis.

% We surprisingly find that in these identities, the proportion of females was relatively high, reaching 64\%, despite the gender balance in the dataset. Similar phenomena have appeared in face recognition, where the accuracy of females is lower than that of males \cite{buolamwini2018gender}. According to our observation, even for similar voices, female's face images are more various and easily affected by factors such as makeup and hairstyle.

\noindent \textbf{Effect of pre-training.} We also explored the effect of pre-training on the model in {\it Supplementary Materials}.
% We also explored the effect of pre-training on the model. We trained three models under different pre-training configurations: pre-train face encoder only, pre-train voice encoder only and without pre-training. From the results shown in \tabref{tab:pt}, it could be found that the pre-training on both encoders is helpful to the model accuracy, and the accuracy drops about 5\% without any pre-training. Comparing the first three variants, merely pre-training face encoder cannot improve the performance, which shows that the main deficiency of the current dataset is the lack of audio data diversity. Generally speaking, pre-training is necessary for the current dataset, which also reflects that there is still room for improvement in the scale and diversity of the current dataset. Larger-scale data, especially the increase of audio data, is helpful for better learning the association between face and voice \cite{xiong2019voice}.

\begin{table}[t]
	\caption{The effect of weighting and personalized identities filtering on 1:2 matching accuracy. The best results are shown in \textbf{bold}.}
    \centering
    \begin{tabular}{c|ccc}
    \toprule
    $R_{keep}$ &
    % \multirow{2}{*}{$R_{keep}$}
    % & \multicolumn{2}{c}{1:2 Matching (ACC\%)}\\
    re-weighting & V-F (U) & F-V (U)\\
    \midrule
    1.0 & $\times$ & 85.5 & 85.2 \\
    1.0 & \checkmark & 86.7 & 86.8 \\
    0.9 & \checkmark & \textbf{87.2} & \textbf{87.2} \\
    0.8 & \checkmark & 87.0 & 87.1 \\
    0.7 & \checkmark & 86.3 & 86.3 \\
    0.6 & \checkmark & 85.9 & 85.3 \\
    \bottomrule
    \end{tabular}
    \label{tab:keep_ratio}
\end{table}

% \begin{table}[t]
% 	\caption{The effect of pre-training on voice and face extractors on 1:2 matching accuracy. The best results are shown in \textbf{bold}.}
%     \centering
%     \begin{tabular}{c|cccc}
%     \toprule No. & Pretrain-F & Pretrain-V & V-F (U) & F-V (U)\\
%     \midrule
%     1 & $\times$ & $\times$ & 82.1 & 82.4 \\
%     2 & \checkmark & $\times$ & 81.8 & 82.4 \\
%     3 & $\times$ & \checkmark & 84.6 & 84.4 \\
%     4 & \checkmark & \checkmark & \textbf{87.2} & \textbf{87.2} \\
%     \bottomrule
%     \end{tabular}
%     \label{tab:pt}
% \end{table}

\section{Conclusion}
In this paper, we propose a novel embedding framework for voice-face assoication learning, with \textbf{(a)} a comprehensive modality alignment loss embracing global and local information;  \textbf{(b)} a dynamic re-weighting strategy to deal with the diversity of learning difficulty across identities.  For  \textbf{(a)}, we propose a two-level loss including implicit modality alignment loss and explicit modality alignment loss. In Prop. \ref{prop:implicit}, we prove that the implicit alignment loss could globally reduce the inconsistency of the embeddings across modality.
% Moreover, we also introduce a explicit alignment loss to complement the global information from a local perspective via pairwise comparisons sampled from each mini-batch.
For \textbf{(b)}, we propose a re-weighting framework which can focus
on hard identities while filter out personalized identities via the evolution of the identity weights. Experiments on cross-modal matching, verification and retrieval show that compared with previous methods, our method can better learn the association between face and voice.

\section*{Acknowledgments}
% \begin{acks}
    This work was supported in part by the National Key R\&D Program of China under Grant 2018AAA0102003, in part by National Natural Science Foundation of China: 61620106009, 61931008, 61836002, and 61976202, in part by Youth Innovation Promotion Association CAS, and in part by the Strategic Priority Research Program of Chinese Academy of Sciences, Grant No. XDB28000000.
% \end{acks}

{\small
\balance
\bibliographystyle{ieee_fullname}
\bibliography{cvpr}

\begin{thebibliography}{10}\itemsep=-1pt

\bibitem{arandjelovic2016netvlad}
Relja Arandjelovic, Petr Gronat, Akihiko Torii, Tomas Pajdla, and Josef Sivic.
\newblock Netvlad: Cnn architecture for weakly supervised place recognition.
\newblock In {\em IEEE Conference on Computer Vision and Pattern Recognition},
  pages 5297--5307, 2016.

\bibitem{bai2020speech}
Yeqi Bai, Tao Ma, Lipo Wang, and Zhenjie Zhang.
\newblock Speech fusion to face: Bridging the gap between human's vocal
  characteristics and facial imaging.
\newblock {\em arXiv preprint arXiv:2006.05888}, 2020.

\bibitem{choi2019inference}
Hyeong-Seok Choi, Changdae Park, and Kyogu Lee.
\newblock From inference to generation: End-to-end fully self-supervised
  generation of human face from speech.
\newblock In {\em International Conference on Learning Representations}, 2019.

\bibitem{chung2018voxceleb2}
Joon~Son Chung, Arsha Nagrani, and Andrew Zisserman.
\newblock Voxceleb2: Deep speaker recognition.
\newblock {\em Proc. Interspeech}, pages 1086--1090, 2018.

\bibitem{fan2017learning}
Yang Fan, Fei Tian, Tao Qin, Jiang Bian, and Tie-Yan Liu.
\newblock Learning what data to learn.
\newblock {\em arXiv preprint arXiv:1702.08635}, 2017.

\bibitem{guo2016ms}
Yandong Guo, Lei Zhang, Yuxiao Hu, Xiaodong He, and Jianfeng Gao.
\newblock Ms-celeb-1m: A dataset and benchmark for large-scale face
  recognition.
\newblock In {\em European Conference on Computer Vision}, pages 87--102, 2016.

\bibitem{hadsell2006dimensionality}
Raia Hadsell, Sumit Chopra, and Yann LeCun.
\newblock Dimensionality reduction by learning an invariant mapping.
\newblock In {\em IEEE Computer Society Conference on Computer Vision and
  Pattern Recognition}, volume~2, pages 1735--1742, 2006.

\bibitem{horiguchi2018face}
Shota Horiguchi, Naoyuki Kanda, and Kenji Nagamatsu.
\newblock Face-voice matching using cross-modal embeddings.
\newblock In {\em ACM International Conference on Multimedia}, pages
  1011--1019, 2018.

\bibitem{hu2018squeeze}
Jie Hu, Li Shen, and Gang Sun.
\newblock Squeeze-and-excitation networks.
\newblock In {\em IEEE conference on computer vision and pattern recognition},
  pages 7132--7141, 2018.

\bibitem{kamachi2003putting}
Miyuki Kamachi, Harold Hill, Karen Lander, and Eric Vatikiotis-Bateson.
\newblock Putting the face to the voice': Matching identity across modality.
\newblock {\em Current Biology}, 13(19):1709--1714, 2003.

\bibitem{kim2018learning}
Changil Kim, Hijung~Valentina Shin, Tae-Hyun Oh, Alexandre Kaspar, Mohamed
  Elgharib, and Wojciech Matusik.
\newblock On learning associations of faces and voices.
\newblock In {\em Asian Conference on Computer Vision}, pages 276--292, 2018.

\bibitem{koh2017understanding}
Pang~Wei Koh and Percy Liang.
\newblock Understanding black-box predictions via influence functions.
\newblock In {\em International Conference on Machine Learning}, pages
  1885--1894, 2017.

\bibitem{li2019gradient}
Buyu Li, Yu Liu, and Xiaogang Wang.
\newblock Gradient harmonized single-stage detector.
\newblock In {\em AAAI Conference on Artificial Intelligence}, volume~33, pages
  8577--8584, 2019.

\bibitem{lin2017focal}
Tsung-Yi Lin, Priya Goyal, Ross Girshick, Kaiming He, and Piotr Doll{\'a}r.
\newblock Focal loss for dense object detection.
\newblock In {\em IEEE International Conference on Computer Vision}, pages
  2980--2988, 2017.

\bibitem{mavica2013matching}
Lauren~W Mavica and Elan Barenholtz.
\newblock Matching voice and face identity from static images.
\newblock {\em Journal of Experimental Psychology: Human Perception and
  Performance}, 39(2):307, 2013.

\bibitem{nagrani2018learnable}
Arsha Nagrani, Samuel Albanie, and Andrew Zisserman.
\newblock Learnable pins: Cross-modal embeddings for person identity.
\newblock In {\em European Conference on Computer Vision}, pages 71--88, 2018.

\bibitem{nagrani2018seeing}
Arsha Nagrani, Samuel Albanie, and Andrew Zisserman.
\newblock Seeing voices and hearing faces: Cross-modal biometric matching.
\newblock In {\em IEEE Conference on Computer Vision and Pattern Recognition},
  pages 8427--8436, 2018.

\bibitem{nagrani2017voxceleb}
Arsha Nagrani, Joon~Son Chung, and Andrew Zisserman.
\newblock Voxceleb: a large-scale speaker identification dataset.
\newblock {\em Telephony}, 3:33--039, 2017.

\bibitem{oh2019speech2face}
Tae-Hyun Oh, Tali Dekel, Changil Kim, Inbar Mosseri, William~T Freeman, Michael
  Rubinstein, and Wojciech Matusik.
\newblock Speech2face: Learning the face behind a voice.
\newblock In {\em IEEE Conference on Computer Vision and Pattern Recognition},
  pages 7539--7548, 2019.

\bibitem{parkhi2015deep}
Omkar~M. Parkhi, Andrea Vedaldi, and Andrew Zisserman.
\newblock Deep face recognition.
\newblock In {\em British Machine Vision Conference}, pages 41.1--41.12,
  September 2015.

\bibitem{ren2018learning}
Mengye Ren, Wenyuan Zeng, Bin Yang, and Raquel Urtasun.
\newblock Learning to reweight examples for robust deep learning.
\newblock In {\em International Conference on Machine Learning}, pages
  4334--4343, 2018.

\bibitem{shrivastava2016training}
Abhinav Shrivastava, Abhinav Gupta, and Ross Girshick.
\newblock Training region-based object detectors with online hard example
  mining.
\newblock In {\em IEEE Conference on Computer Vision and Pattern Recognition},
  pages 761--769, 2016.

\bibitem{smith2016matching}
Harriet~MJ Smith, Andrew~K Dunn, Thom Baguley, and Paula~C Stacey.
\newblock Matching novel face and voice identity using static and dynamic
  facial images.
\newblock {\em Attention, Perception, \& Psychophysics}, 78(3):868--879, 2016.

\bibitem{sohn2016improved}
Kihyuk Sohn.
\newblock Improved deep metric learning with multi-class n-pair loss objective.
\newblock In {\em Advances in Neural Information Processing Systems}, pages
  1857--1865, 2016.

\bibitem{sun2020circle}
Yifan Sun, Changmao Cheng, Yuhan Zhang, Chi Zhang, Liang Zheng, Zhongdao Wang,
  and Yichen Wei.
\newblock Circle loss: A unified perspective of pair similarity optimization.
\newblock In {\em IEEE/CVF Conference on Computer Vision and Pattern
  Recognition}, pages 6398--6407, 2020.

\bibitem{tsantani2019brain}
Maria~Stephanie Tsantani.
\newblock {\em Brain and perceptual representations of faces, voices, and
  person identity}.
\newblock PhD thesis, Brunel University London, 2019.

\bibitem{wang2020learning}
Rui Wang, Xin Liu, Yiu-ming Cheung, Kai Cheng, Nannan Wang, and Wentao Fan.
\newblock Learning discriminative joint embeddings for efficient face and voice
  association.
\newblock In {\em International ACM SIGIR Conference on Research and
  Development in Information Retrieval}, pages 1881--1884, 2020.

\bibitem{wang2018data}
Tianyang Wang, Jun Huan, and Bo Li.
\newblock Data dropout: Optimizing training data for convolutional neural
  networks.
\newblock In {\em IEEE International Conference on Tools with Artificial
  Intelligence}, pages 39--46, 2018.

\bibitem{weinberger2006distance}
Kilian~Q Weinberger, John Blitzer, and Lawrence~K Saul.
\newblock Distance metric learning for large margin nearest neighbor
  classification.
\newblock In {\em Advances in Neural Information Processing Systems}, pages
  1473--1480, 2006.

\bibitem{wen2018disjoint}
Yandong Wen, Mahmoud Al~Ismail, Weiyang Liu, Bhiksha Raj, and Rita Singh.
\newblock Disjoint mapping network for cross-modal matching of voices and
  faces.
\newblock In {\em International Conference on Learning Representations}, 2018.

\bibitem{wen2019face}
Yandong Wen, Bhiksha Raj, and Rita Singh.
\newblock Face reconstruction from voice using generative adversarial networks.
\newblock In {\em Advances in Neural Information Processing Systems}, pages
  5265--5274, 2019.

\bibitem{wickelmaier2003introduction}
Florian Wickelmaier.
\newblock An introduction to mds.
\newblock {\em Sound Quality Research Unit, Aalborg University, Denmark},
  46(5):1--26, 2003.

\bibitem{xiong2019voice}
Chuyuan Xiong, Deyuan Zhang, Tao Liu, and Xiaoyong Du.
\newblock Voice-face cross-modal matching and retrieval: A benchmark.
\newblock {\em arXiv preprint arXiv:1911.09338}, 2019.

\bibitem{zhang2020can}
Y Zhang, S Yang, J Xiao, S Shan, and X Chen.
\newblock Can we read speech beyond the lips? rethinking roi selection for deep
  visual speech recognition.
\newblock In {\em IEEE International Conference on Automatic Face and Gesture
  Recognition}, pages 851--858, 2020.

\bibitem{zhong2018ghostvlad}
Yujie Zhong, Relja Arandjelovi{\'c}, and Andrew Zisserman.
\newblock Ghostvlad for set-based face recognition.
\newblock In {\em Asian Conference on Computer Vision}, pages 35--50, 2018.

\end{thebibliography}
}

\end{document}